\pdfoutput=1

\documentclass[11pt]{article}

\usepackage[final]{acl}

\usepackage{times}
\usepackage{latexsym}
\usepackage{todonotes}
\usepackage[T1]{fontenc}

\usepackage[utf8]{inputenc}

\usepackage{microtype}

\usepackage{inconsolata}

\usepackage{graphicx}

\usepackage{tabularx}
\newcommand{\mycomment}[1]{}

\title{A Survey of Idiom Datasets \\ for Psycholinguistic and Computational Research}

\author{Michael Flor \\
  Educational Testing Service\\ Princeton, New Jersey, USA \\
\texttt{mflor@ets.org} \\\And
   Xinyi Liu \\
    Montclair State University \\ Montclair, New Jersey, USA\\
    \texttt{liux2@montclair.edu} 
   \AND Anna Feldman\\
   Montclair State University, Montclair, New Jersey, USA\\
  \texttt{feldmana@montclair.edu} \\
  }

\begin{document}
\maketitle
\begin{abstract}
Idioms are figurative expressions whose meanings often cannot be inferred from their individual words, making them difficult to process computationally and posing challenges for human experimental studies. This survey reviews datasets developed in psycholinguistics and computational linguistics for studying idioms, focusing on their content, form, and intended use. Psycholinguistic resources typically contain normed ratings along dimensions such as familiarity, transparency, and compositionality, while computational datasets support tasks like idiomaticity detection/classification,  paraphrasing, and cross-lingual modeling.  We present trends in annotation practices, coverage, and task framing across 53 datasets.  Although recent efforts expanded language coverage and task diversity, there seems to be no relation yet between psycholinguistic and computational research on idioms. 

\end{abstract}

\section{Introduction}
Idioms are conventional expressions whose meanings cannot be reliably inferred from the meanings of their individual words. Phrases such as \textit{kick the bucket} or \textit{spill the beans} demonstrate how idioms often convey figurative meanings that diverge from literal interpretations. Because of their semantic opacity and syntactic variability, idioms present persistent challenges for natural language processing (NLP) systems.

Recognizing and interpreting idiomatic expressions is essential for a range of NLP applications, including machine translation, dialogue systems, and sentiment analysis. However, automatic idiom processing remains difficult. The same expression can be used literally or figuratively depending on context, and idioms vary in their degree of flexibility, compositionality, and transparency. Addressing these challenges requires high-quality datasets that capture the complexities of idiomatic language in realistic contexts.

Over the past two decades, a number of datasets for idiom processing have been created, often as part of individual research projects. However, these resources are highly heterogeneous: they differ in their annotation schemes, languages covered, idiom types included, and dataset sizes. There is currently no standard benchmark, no unified annotation framework, and limited cross-dataset compatibility, making it difficult to compare methods or advance the field systematically.

This paper surveys the available datasets for idiom analysis and processing. Those come from two different disciplines - psycholinguistics and computational linguistics. We provide an overview of their properties, including language coverage, annotation strategies, intended tasks, and dataset sizes. We highlight the strengths and limitations of existing resources and identify areas where further development is needed to support progress in idiom-aware NLP.  A continuously updated list of idiom datasets, with links to datasets and publications, is maintained at \textit{\url{https://github.com/maafiah/IdiomsResearch}}.




\section{Idiom datasets in psycholinguistic research}

In psycholinguistics, research on idioms has been largely focused on the immediate
 processing of idioms, such as reading time, most notably the comprehension of idioms in (and out of) context. 
The motivations for specific hypotheses in such research are often related to the linguistically-informed theories or models of idiom processing \cite{Cacciari2014,EspinalMateu2019}. 

The time-course of idiom processing by human subjects is influenced by a variety of variables, such as idiom familiarity. 
Controlling for such variables is an important feature of psycholinguistic experiments. 
Some of those variables are motivated by linguistic theories, other variables are motivated by psychological considerations. 
Norming studies allow researchers to collect subjective individual ratings for a variety of postulated dimensions and aggregate them across respondents \cite{Winter2022}. Idiom norming studies use Likert scales to rate idiomatic expressions on various dimensions. However, the constructs are not always the same across studies. Table \ref{PL-datasets} in the appendix lists 16 public datasets dedicated to psycholinguistic aspects of idiomatic expressions. Most of those datasets are publicly available. The aspects/dimensions are described below.

\vspace{1mm}
\textbf{Familiarity} of an idiom has been construed as the frequency with which a person has been exposed to a given expression in their everyday life. However, since such frequency cannot be measured objectively, the standard approach is to ask raters for a subjective estimation (\textit{subjective frequency}), explained as ‘familiarity’.
Several studies demonstrated that familiar idioms are processed faster than less familiar ones  \citep{Schweigert1986, SchweigertMoates1988}.

\vspace{1mm}
\textbf{Knowledge of meaning}. A notion related to familiarity is knowledge, the degree to which a person thinks they know the meaning of an idiom, and can explain what the expression means \cite{LiZhangWang2016}. This also sometimes serves as a control on the adequacy of participant ratings \cite{Pagliai2023}.

The notion of \textbf{literality} (also called literalness or ambiguity) concerns whether the idiom has a plausible literal interpretation. 
For example, \textit{break the ice} has a literal interpretation, but \textit{shoot the breeze} is semantically anomalous because a breeze is not something that can be shot. 
Literality is usually measured by asking
participants to rate whether the phrase could be used literally in addition to its figurative meaning \citep{LibbenTitone2008, Caillies2009, TabossiETAL2011}. While literality could be conceived as a dichotomous variable, it is usually measured on a scale.

\vspace{1mm}
\textbf{Compositionality} and \textbf{decomposability}. 
This dimension estimates the degree to which the idiom's component words contribute to its idiomatic interpretation.
This is based on the notion that some idioms
can be compositionally analyzed \cite{NSW1994}.
For example, for \textit{spill the beans}, meaning 'reveal secrets', the verb 'spill' corresponds to 'reveal', 'beans' corresponds to 'secrets', showing some composition of the whole figurative meaning. No such composition occurs in e.g. \textit{kick the bucket}, that idiom is not decomposable. 
\citet{HamblinGibbs1999} suggested that the degree to which the meaning of idiom parts contributes to idiom interpretation can be classified on a continuum, calling it \textit{degree of analyzability}. 
Experimental studies have found mixed effects of idiom analyzability on processing time 
\cite{TitoneLibben2014, TabossiFanariWolf2008}.

\vspace{1mm}
\textbf{Transparency} is defined as the ease by which the motivation for the structure of an idiomatic expression can be deduced \cite{NSW1994}, or how easy it is to recognize why  an idiom means what it means. Such motivation can be a metaphorical relation (e.g., the idiom \textit{flip one's lid} might be based on a metaphor that anger is like steam), or a historical remnant (e.g., '\textit{carry coals to Newcastle}'), etc. The reciprocal notion for transparency is \textit{opacity}. Semantic (or conceptual) transparency of idioms is not the same notion as compositionality, but they are related as the ratings are often highly correlated. According to \citet{CitronETAL2016}, transparency is a problematic  measure since participant ratings are based on intuitions and guesses, and highly dependent on the knowledge of the correct idiomatic meaning. 

\vspace{1mm}
\textbf{Age of acquisition} (AoA) is the estimated age at which 
a word (or expression) and its meaning were first learned \citep{Johnston01052006}. AoA is typically measured in years and months. While AoA for words has been a common measure in psycholinguistics \cite{aoa1997}, norming AoA estimations for idiomatic expressions was introduced by \citet{TabossiETAL2011}. 
AoA was found to correlate with knowledge, familiarity, and subjective frequency for idioms \cite{BoninETAL2013,BoninETAL2018, LiZhangWang2016}.  \citet{LiZhangWang2016} suggested that the earlier an idiom is learned, the more frequently it might be encountered, and thus  become more familiar. \citet{BoninETAL2013} found that AoA was predictive for idiom reading times. \citet{BoninETAL2018} suggested that AoA norms might be useful for studying idiom processing in children.

\vspace{1mm}
\textbf{Predictability} of idioms is defined as the probability of completing an incomplete string to a full idiomatic expression, for example completing \textit{'be in seventh....'} with \textit{'heaven'}. For measuring predictability, participants are asked to read incomplete sentences with idioms and provide the last word of the idiom (that is blanked out).

\vspace{1mm}
\textbf{Syntactic flexibility} refers to the notion that idiomatic expressions are often not entirely frozen and allow some degrees of syntactic variability, such as inflection of verbs, insertions of adjective or adverbs, sometimes passivization, etc., \citep{Fraser1970, Moon1998}.
As noted by \citet{GibbsGonzales1985}, this can have psychological implications. For example, flexible idioms can easily undergo
transformations, e.g., for \textit{throw in the towel}, the passive form \textit{the towel was thrown in by him}  retains the idiomatic meaning \textit{he gave up}. Less flexible expressions are less likely to be interpreted figuratively, e.g., \textit{the bucket was kicked by him} is less likely to be
interpreted as \textit{he died}. For norming studies, idiomatic expressions are presented in various forms and participants are asked to rate their idiomaticity \cite{TabossiETAL2011}.  

The notion of \textbf{concreteness} describes the extent to which the meaning of a word  refers to a state or event that can be experienced via one or more sensory modalities \citep{PaivioETAL1968}.
In experiments, participants process concrete words faster and more accurately than abstract words (the so called \textit{concreteness effect}, \citealt{Paivio1991}). Two studies collected concreteness ratings for idiomatic phrases, as opposed to concreteness of constituent words \citep{CitronETAL2016,Morid_Sabourin_2024}. Whether concreteness of idioms might be related to idiom processing or representation is yet unknown.

\vspace{1mm}
\textbf{Imageability} refers to the ability to create a mental image of a word. Generally, imageability enhances word recognition \citep{CONNELL2012452}. Imageability is often highly correlated with concreteness. Mental imagery  has been linked to idiom comprehension \citep{GibbsOBrien1990}, and norming imageability of idioms is a new research trend.

In psycholinguistics, three important non-cognitive aspects are known to have influence on representation and meaning of individual words \citep{OsgoodETAL, Russell2003}, known collectively as VAD: 
\textbf{valence} (affective value, sentiment, positiveness–negativeness, or pleasantness of a stimulus),
\textbf{arousal} (feeling active or passive, or intensity associated with the stimulus), 
and \textbf{dominance}
(dominant–submissive, degree of control).
VAD norms for individual words are well known in psycholinguistic research (e.g. \citealt{WarrinerETAL2013}), and also in computational linguistics research
\citep{Mohammad2018}. 
\citet{NSW1994} stated that “idioms are typically used to imply a certain evaluation or affective stance toward the things they denote". \citet{PFV2014} used the rated valence of idiom component words for automated detection of idioms in texts. Obtaining valence and arousal ratings of whole idiomatic expressions is a recent trend in norming studies \citep{GavilanETAL2021, Morid_Sabourin_2024}.

\section{Idiom datasets in computational linguistic research}
Computational linguistics research on idioms has produced a wide range of datasets designed for classification, disambiguation, paraphrasing, and multilingual modeling. Unlike psycholinguistic norming studies, which focus on controlled ratings of idiom properties, these resources are typically drawn from real-world corpora and are suited for NLP tasks. Table \ref{table:CL-datasets} in the appendix lists the datasets published in computational-linguistics literature. Most of those datasets are publicly available.

\subsection{Classification and Disambiguation}

A significant portion of computational idiom datasets focus on binary classification -- distinguishing idiomatic from literal uses of the same expression in context. The VNC-Tokens dataset \citep{fazly-etal-2009-unsupervised} includes nearly 3,000 usages of verb-noun combinations in English, annotated as idiomatic or literal. Similarly, the IDIX corpus \citep{IDIX2010} collected over 5,800 English sentences labeled for idiomaticity, though it is not publicly available.

Several benchmark datasets have been created in the context of shared tasks. SemEval-2013 Task 5b \citep{Semeval2013Task5} presents 85 ambiguous idioms with over 4,000 instances, each provided with a five-sentence context for disambiguation. The RU Idioms dataset \citep{aharodnik-etal-2018} includes 2,420 idiomatic instances and 3,027 literal ones, drawn from Classical and Modern Russian literature and Wikipedia texts. Designed to support supervised classification of idiomaticity in Russian, the corpus provides richly contextualized examples in paragraph-level texts.
The MAGPIE dataset \cite{magpie2020} introduced 56,622 instances of potential idioms in short textual context, for 1,756 different English idioms. 

\subsection{Paraphrase, Sentiment, and Substitution}

Beyond disambiguation, idiom paraphrasing has been a central task. \citet{pershina-etal-2015-idiom} collected over 2,400 English idioms with associated paraphrases and 1,400 idiom-idiom pairs annotated for mutual paraphrasability. The Idiom Substitution dataset \citep{liu-hwa-2016-phrasal} offers definitions and plausible substitutions for 172 English idioms, useful for generation tasks. The Parallel Idioms Corpus \cite{zhou-etal-2021-pie} presented 823 English-language idioms with 5,170 sentence-pairs, where a sentence contains an idiom or its literal paraphrase.

Sentiment-oriented datasets such as IDIOMENT \citep{WILLIAMS20157375} and SLIDE \citep{jochim-etal-2018-slide} capture affective dimensions. IDIOMENT includes 580 idioms with both idiomatic and sentence-level sentiment annotations, while SLIDE contains sentiment scores for over 5,000 idioms. IDEM \citep{prochnow-etal-2024-idem} has about 9685 sentences with idioms and labels for expressed emotion in each sentence. These resources are especially relevant for tasks like figurative sentiment classification and sentiment-aware generation.

\subsection{Multilingual Resources}

Recent efforts have addressed the lack of multilingual idiom data. The LIdioms dataset \citep{moussallem-etal-2018-lidioms} provides 815 idioms across five languages (English, German, Italian, Portuguese, and Russian) in RDF format, linking semantically equivalent expressions. The IMIL corpus \citep{agrawal-etal-2018-beating} contains over 2,200 English idioms and their translations in seven Indian languages, annotated across 250K sentences. PETCI \citep{tang2022petci} includes 4,310 Chinese idioms with 29,936 English translations, capturing diverse translation errors and paraphrase strategies.

SemEval-2022 Task 2 \citep{tayyar-madabushi-etal-2022-semeval} extended idiomaticity detection to three languages -- English, Portuguese, and Galician -- by providing labeled training and development data for English and Portuguese, and zero-shot test data for Galician. While this marked a step toward multilingual idiom processing, the dataset does not include aligned idiom instances across languages or annotations for contextual factors such as register, familiarity, or cultural variation, limiting its utility for studying pragmatic differences.

IdiomKB \cite{IdiomKB2024} merged several previously published idiom datasets (for English, Chinese, and Japanese), with the purpose of improving cross-lingual LLM-based translation of texts with idiomatic expressions.

\subsection{Model Probing and Representation Learning}

Several datasets have been developed to probe and improve language models' handling of idioms. AStitchInLanguageModels \citep{tayyar-madabushi-etal-2021-astitchinlanguagemodels-dataset} comprises naturally occurring sentences containing potentially idiomatic multiword expressions (MWEs) in English and Portuguese, annotated with fine-grained meanings and paraphrases. This dataset supports tasks evaluating models' ability to detect idiom usage and generate effective representations of idiomaticity.

IDIOMEM \citep{haviv-etal-2023-understanding} is a probe dataset of English idioms used to analyze memorization behavior in transformer language models. It facilitates the study of when and how models recall memorized idiomatic sequences. CultureLLM \citep{culturellm2024} focuses on incorporating cultural differences into large language models (LLMs),  generating semantically equivalent training data through semantic data augmentation, fine-tuning culture-specific LLMs for nine cultures. \citet{LiuKotoETAL2024} and \citet{khoshtab2025comparativestudymultilingualidioms} use proverbs and idioms to probe LLMs inference in processing figurative language across multiple languages.

\subsection{Gaps and Future Directions}

While these datasets have enabled significant progress in idiom-aware NLP, several limitations persist. Annotation schemes vary widely across datasets—some mark idiomaticity at the phrase level, others at the sentence level; some provide paraphrases or sentiment labels, others do not. This heterogeneity reflects the diversity of research goals, but it also makes cross-dataset evaluation difficult. Dataset sizes and scopes differ considerably: some focus on depth (many instances per idiom), others on breadth (many idioms with few examples). While multilingual coverage is growing, most datasets do not include semantically aligned idioms across languages or cultural contexts. In addition, idioms are drawn from limited domains, with little attention to genre, register, or discourse variation. Rather than imposing a single standard, future work could benefit from shared metadata conventions, interoperable formats, and more transparent documentation to support comparison, reuse, and integration across datasets.
In addition, there seems to be no relation yet between psycholinguistic and computational research on idioms. This is not surprising, as those research domains traditionally had different orientations and research agendas. Notably, datasets from psycholinguistics focus on idioms as 'types', while computational datasets often work with idioms as 'tokens' (instances) in various textual contexts. One area where some cross-pollination between the fields may develop is the notion of valence (sentiment) of idioms. Another possible direction could be using computational methods to model/explain human ratings on various aspects of idioms. 

\section{Conclusion}
This survey compares idioms datasets developed in psycholinguistics and computational linguistics, focusing on their design, intended use, and underlying assumptions. By examining both types, we highlight how methodological differences shape what each dataset can support. 

\section*{Limitations}
While we conducted an extensive search for various research-oriented datasets of idiomatic expressions, 
some resources might have been missed. 

\section*{Acknowledgments}
This material is partially based upon work supported by the U.S. National Science Foundation under Grant Numbers 2428506 and 2226006.

\bibliography{custom}

\appendix

\section{Appendix}
\label{sec:appendix}
In this appendix we list datasets from psycholinguistic research in Table 1, and datasets from computational linguistics research in Table 2.

\newcolumntype{L}[1]{>{\raggedright\arraybackslash}p{#1}} 

\begin{table*}
  \centering
\begin{tabularx}{\textwidth}{p{3.1cm}llp{7cm}c}
%
    \hline
    \textbf{Authors} & \textbf{Count} & \textbf{Language} & \textbf{Contents (rating dimensions)} & Avail. \\
    \hline
    \citet{LibbenTitone2008} & 210 & English  & familiarity, meaningfulness, literal plausibility, decomposability   & y    \\
    \hline
    \citet{Caillies2009} & 300 & French & familiarity, knowledge of  meaning, literality, compositionality and
predictability & y 
    \\
    \hline
    \citet{TabossiETAL2011} & 245 & Italian & knowledge, familiarity, AoA, predictability, syntactic flexibility, compositionality & y
    \\
    \hline
    \citet{BoninETAL2013}       & 305           & French &knowledge, familiarity, subjective and objective frequency, AoA, predictability, literality, compositionality, and length & y 
    \\
    \hline
    \citet{CitronETAL2016} & 619 & German & emotional valence, arousal, familiarity, semantic transparency, figurativeness, concreteness & y
    \\
    \hline
    \citet{BeckWeber2016} & 300 & German & meaningfulness, familiarity, literality, decomposability; & y
    \\
    \citet{Beck2020} &  & English & ratings by L1 and L2 speakers & 
    \\
    \hline
    \citet{LiZhangWang2016} & 350 & Chinese & knowledge, familiarity, subjective frequency, AoA, predictability, literality, compositionality & y
    \\
    \hline
    \citet{BulkesTanner2017}    &  870  & English &
    familiarity, meaningfulness, literal plausibility, decomposability, predictability & y 
    \\
    \hline
    \citet{NordmanJambazova2017} & 90 & Bulgarian & familiarity, compositionality, literality, etc. & y \\
       & 100 & English 
       \\
       \hline
    \citet{BoninETAL2018} & 160+  & French & knowledge, predictability, literality, compositionality, subjective and objective frequency, familiarity, AoA, length & y              \\
    \hline
    \citet{HubersETAL2019} & 374 & Dutch & frequency of exposure, meaning familiarity, frequency of usage, transparency,  imageability & y 
    \\
    \hline
    \citet{GavilanETAL2021} & 1252 & Spanish & familiarity, knowledge, decomposability, literality, predictability, valence, arousal & y
    \\
    \hline
    \citet{GondoraETAL2022} & 1082 & Chilean-Spanish &  familiarity, ambiguity, compositionality, transparency & n
    \\
    \hline
    \citet{Pagliai2023} & 150 & Italian & familiarity, literality, decomposability, transparency, objective knowledge, meaningfulness & y \\
    & 150 & English 
    \\
    \hline
    \citet{LadaETAL2024} & 400 & Greek & subjective frequency, ambiguity, decomposability & y
    \\
    \hline
    \citet{Morid_Sabourin_2024} & 210 & English & arousal, valence, concreteness, imageability - from English L1 and L2 speakers & y \\
    \hline
\end{tabularx}
\caption{
    Published datasets from psycholinguistic research, listed chronologically. The 'Avail.' column indicates dataset availability. 'Count' indicates number of idioms.}
\label{PL-datasets}
\end{table*}

\begin{table*}
  \centering
\begin{tabularx}{\textwidth}{p{2cm}p{2.8cm}p{1.8cm}p{7cm}c}
    \hline
    \textbf{Dataset} & \textbf{Authors} & \textbf{Language} & \textbf{Contents} & Avail.
    \\
    \hline
    Phraseo-Lex & \citet{DormeyerFischer1998} & German &  verbal idioms (types), with linguistic annotations & n
    \\
     \hline
     Berlin Idiom Project & \citet{BIP2005} & German & 500 idioms (types) with linguistic annotations & n
     \\
     \hline
     VNC-Tokens & \citet{VNC2008} \newline \citet{fazly-etal-2009-unsupervised} & English & 53 types, almost 3000 English verb-noun combination instances annotated as to whether they are literal or idiomatic & y
     \\
    \hline
    OpenMWE & \citet{hashimoto-kawahara-2008} & Japanese & 146 ambiguous idioms (types),  102846 sentences, annotated literal/idiomatic & y
    \\
    \hline
    IDIX & \citet{IDIX2010} & English & 78 idioms (types), 5836 instances annotated as y/n idiomatic & n
    \\
    \hline
    CIKB & \citet{wang-yu-2010-construction} & Chinese & 38K idioms (types) with linguistic annotations & n
    \\
    \hline
    Reddy et al. 2011 & \citet{reddy-etal-2011-empirical} & English & 90 nominal compounds with compositionality ratings & y
    \\
    \hline
    Semeval2013 Task 5b & \citet{Semeval2013Task5} & English & 85 ambiguous idioms (types), 4350 instances, each in 5-sentences context, marked y/n idiomatic & y
    \\
    \hline
    IDIOMENT & \citet{WILLIAMS20157375} & English & 580 idioms (types) with sentiment ratings; 2521 instances in sentence context, with sentiment ratings & y
    \\
    \hline
    Idiom \newline Paraphrases & \citet{pershina-etal-2015-idiom} & English & 2432 idioms (types) with paraphrases, 1400 pairs of idioms annotated as y/n mutual paraphrases  & y
    \\
    \hline
    Idiom \newline substitution & \citet{liu-hwa-2016-phrasal} & English & 172 idioms with definitions and substitution phrases & y
    \\
    \hline
    IMIL & \citet{agrawal-etal-2018-beating} & & 2208 English idioms in English with their translations in seven Indian languages, 250,815 sentences with idioms &  y
    \\
    \hline
    Idiom Translation DS & \citet{fadaee-etal-2018-examining} & English, German & 1500 parallel sentences whose German side contains an idiom,  and 1500 parallel sentences whose English side contains an idiom & y
    \\
    \hline
    LIdioms & \citet{moussallem-etal-2018-lidioms} & English +4 & 815 idioms total, in English, German, Italian, Portuguese, and Russian, with links between idioms across languages (RDF format) & y
    \\
    \hline
    RU idioms & \citet{aharodnik-etal-2018} & Russian & 5.4K instances of 100 idiomatic expressions (3K literal, 2.4K idiomatic), each in paragraph context. & y
   \\
   \hline
   CCT & \citet{jiang-etal-2018-chengyu} & Chinese & 7395 Chengyu form idioms (types)  and 100K context sentences& y
    \\
    \hline
    SLIDE & \citet{jochim-etal-2018-slide} & English & 5000 idioms (types) with sentiment annotations & y
    \\
    \hline
    Senaldi 2019 & \citet{Senaldi2019} & Italian & 90 verb-noun and 24 adjective-noun expressions (types) & y
    \\
    \hline
    Composition. of Nominal Compounds & \citet{cordeiro-etal-2019-unsupervised} & English, French, Portuguese & 190/180/180 nominal compounds (types), rated for compositionality & y
    \\
    \hline
\end{tabularx}
\caption{
    Published datasets from computational linguistics research, listed chronologically. The 'Avail.' column indicates dataset availability. Table continues on the next page.}
\label{table:CL-datasets}
\end{table*}

\begin{table*}
  \centering
\begin{tabularx}{\textwidth}{p{2cm}p{2.8cm}p{1.8cm}p{7cm}c}
    \hline
    \textbf{Dataset} & \textbf{Authors} & \textbf{Language} & \textbf{Contents} & Avail. \\
    \hline
    ChID & \citet{zheng-etal-2019-chid} & Chinese & 3848 Chengyu form idioms (types) and 518K context paragraphs with cloze blanks and multiple options & y
    \\
    \hline
    Swedish MWEs & \citet{kurfali-etal-2020-multi} & Swedish & 96 Swedish multi-word expressions (types), annotated with degree of compositionality & y
    \\
    \hline
    MAGPIE & \citet{magpie2020} &  English  & 56622 instances of idioms in short textual context (1756 different types)
   & y    
   \\
    \hline
    NCS & \citet{garcia-etal-2021-probing} & English, Portuguese & 280 and 180 noun compounds (NCs) in English and Portuguese, with 5620/3600 sentences, marked for compositionality & y 
    \\
    \hline
    EPIE &  \citet{EPIE2021} & English & 21891 static idiom instances in sentence context (359 types) and 3135 formal idiom instances in sentence context (358 types) & y
    \\
    \hline
    PIE & \citet{zhou-etal-2021-pie} & English & 823 idioms (types) with 5170 sentence-pairs  containing those idioms or their literal paraphrases. & y 
    \\
    \hline
   AStitchIn LanguageModels & \citet{tayyar-madabushi-etal-2021-astitchinlanguagemodels-dataset} & English \newline Portuguese & 223/113 nominal compounds in sentence context (4558/1872 instances), annotated as literal, idiomatic, proper noun, or ‘meta usage’ & y 
    \\
    \hline
    Semeval2022 Task 2 & \citet{tayyar-madabushi-etal-2022-semeval} & English, Portuguese, Galician &  
    5352/2555/776 instances in sentence context, extends the 
AStitchInLanguageModels dataset & y 
    \\
    \hline
    PIE-English & \citet{PIE_Eng2022} & English & 20100 samples with almost 1,200 cases of idioms (with their meanings), classified into 10 types of figurative expressions & y 
    \\
    \hline
    FLUTE & \citet{chakrabarty-etal-2022-flute} & English & Part of larger dataset on figurative language. Has 1000 idiomatic sentences paired with sentences that entail or contradict the focal sentence & y
    \\
    \hline
    PETCI & \citet{tang2022petci} & Chinese, English & 4310 Chinese idioms with 29,936 English translations.& y 
    \\
    \hline
    ID10M & \citet{tedeschi-etal-2022-id10m} & multiple & Training data in 10 languages was autogenerated: 10K idioms (types) 262781 sentences (32\% idioms); gold test data: only for English, German, Italian Spanish, 200 sentences each & y
    \\
    \hline
    IDIOMEM & \citet{haviv-etal-2023-understanding} & English & 814 idiom types & y
    \\
    \hline
    IDEM & \citet{prochnow-etal-2024-idem} & English & 9685 idiom-containing sentences, labelled with emotions & y
    \\
    \hline
    CultureLLM & \citet{culturellm2024} & & & y
    \\
    \hline
    IdiomKB & \citet{IdiomKB2024} & multiple & A merger of several datasets in English, Chinese and Japanese & y
    \\
    \hline
    MAPS & \cite{LiuKotoETAL2024} & 6 languages & proverbs and sayings,
in English, German, Russian, Bengali,  Chinese, Indonesian (364 to 424 per language); with entailing and non-entailing continuations & y
    \\
    \hline
    Multilingual Idioms and Similes in LLMs & \citet{khoshtab2025comparativestudymultilingualidioms} & Persian, \newline English \newline (11 total) & 316 instances of idioms in LLM-generated  sentence context, with entailing and non-entailing continuations. English translations provided. Also used previous figurative language datasets & y \\
    \hline
\end{tabularx}
\renewcommand\thetable{2} 
\caption{
    Continued: Published datasets from computational linguistics research. }
\end{table*}

\end{document}